# Answering Chinese Elementary School Social Study Multiple Choice Questions

Daniel Lee[1]
Dept. of Computer Science Engineering
University of Michigan
Ann Arbor, Michigan, USA
danclee@umich.edu

Chao-Chun Liang
Institute of Information Science,
Academia Sinica
Taipei, Taiwan
ccliang@iis.sinica.edu.tw

Keh-Yih Su
Institute of Information Science,
Academia Sinica
Taipei, Taiwan
kysu@iis.sinica.edu.tw

*Abstract*—We present a novel approach to answer the Chinese elementary school Social Study Multiple Choice questions. Although BERT has demonstrated excellent performance on *Reading Comprehension* tasks, it is found not good at handling some specific types of questions, such as *Negation*, *All-of-the-above*, and *None-of-the-above*. We thus propose a novel framework to cascade BERT with a *Pre-Processor* and an *Answer-Selector* modules to tackle the above challenges. Experimental results show the proposed approach effectively improves the performance of BERT, and thus demonstrate the feasibility of supplementing BERT with additional modules.

*Keywords—natural language inference, machine reading comprehension, multiple choice question, question and answering*

## I. Introduction

Since any *Natural Language Inference* (NLI) related issue can be checked by asking an appropriate corresponding question, the *Question and Answering* (QA) task has become a very suitable testbed for evaluating NLI progress [1], and *Reading Comprehension (RC) Test* is the one that is frequently adopted. There are various forms of RC questions, including *Cloze*, *Multiple Choice* (MC), and *Binary Choice*. In this paper, we focus on solving MC questions of traditional Chinese *Social Studies* subject of primary school. In this **C**hinese **S**ocial **S**tudies **MC** QA task (abbreviated as CSSMC), the system selects the correct answer from several candidate options based on a given question and its associated lesson. Table I shows an example of this CSSMC data-set, and the passage is the corresponding supporting evidence (SE).

Previous works of answering MC questions can be classified into two categories: statistics-based approaches [2, 3], and neural-network-based approaches [4, 5]. Recently, the pre-trained language models (such as BERT [6], XLNET [7], RoBERTa [8], ALBERT [9]) have shown excellent performance on many MCRC tasks. As BERT shows excellent performance on various English data-sets (e.g., SQuAD 1.1 [10], GLUE [11], etc.), it is adopted as our baseline. Table V shows its performance given the gold SE.

After analyzing some error cases, it is observed that BERT is weak in handling following question types: (1) *Negation* questions type, which means the question passage includes negation phrases (such as '不可能 (unlikely)'). For example, BERT selects the same answer for "小敏的媽媽目前在郵局服務，請問小敏的媽媽可能會為居民提供什麼服務？" and "小敏的媽媽目前在郵局服務，請問小敏的媽媽<u>不</u>可能會為居民提供什麼服務？" (with one additional negation word "<u>不</u>"). It seems BERT does not pay special attention to those negative words; however, even one negation word could change the desired answer. (2) *All-of-the-above* ("以上皆是") and *None-of-the-above* ("以上皆非") question types, which indicate that the options include either "All-of-the-above" or "None-of-the-above". These types of questions cannot be handled by simply picking up the most likely choice without pre-processing the given choices. Table II shows an example of each type mentioned above.

TABLE I. AN EXAMPLE OF SOCIAL STUDY MC QUESTION

| | |
|---|---|
| Passage | 三代同堂家庭是子女和父母、祖父母或外祖父母同住。 |
| Question | 「我和爸爸、媽媽、爺爺、奶奶住在一起。」是屬於哪一種類型的家庭？ |
| Options | (1) 三代同堂家庭<br>(2) 單親家庭<br>(3) 隔代教養家庭<br>(4) 寄養家庭 |
| Answer | (1) 三代同堂家庭 |

TABLE II. DIFFERENT QUESTION TYPES IN CSSMC CORPUS

| Problem Type | Questions |
|---|---|
| Negation | **Question**: "浩浩跟家人到臺東縣關山鎮遊玩，他<u>不</u>可能在當地看到什麼？"<br>**Options**: (1)阿美族豐年祭 (2)環鎮自行車道 (3)油桐花婚禮 (4)親水公園 |
| All of the Above | **Question**: "在高齡化的社會裡，我們應該如何因應高齡化社會的到來？"<br>**Options**: (1)制定老人福利政策 (2)提供良好的安養照顧 (3)建立健全的醫療體系 (4)<u>以上皆是</u> |
| None of the Above | **Question**: "都市有公共設施完善、工作機會多等優點，常吸引鄉村地區哪一種年齡層的居民前往？"<br>**Options**: (1)老人 (2)小孩 (3)青壯年 (4)<u>以上皆非</u> |

However, it is difficult to pinpoint the deficit spots that cause the problem and then make the corresponding remedies within BERT, as it has a complicated architecture (with 12 heads and 12 stacked layers). Therefore, we prefer to keep its implementation untouched, if the problem could be fixed by

---

coupling BERT with some external modules. We thus propose a framework which cascades BERT with a *Pre-Processor* module, which is responsible for handling the question-types of *All-of-the-above* and *None-of-the-above*, and an *Answer-Selector* module (a post-processor), which is responsible for handling the negation question-type.

The *Answer-Selector* module selects the candidate with the lowest score, instead of the highest one as what we usually adopt, when the question type is *Negation*. On the other hand, the *Pre-processor* module replaces either "*All-of-the-above*" or "*None-of-the-above*" original choice with a new choice which is generated by concatenating all other choices together. Take the second last row in Table II as an example, we will replace the original last choice "以上皆是" with "制定老人福利政策^提供良好的安養照顧^建立健全的醫療體系".

We test the proposed framework on the CSSMC data-set. The experimental results show the proposed approaches outperform the pure BERT model. It thus shows a new way to supplement BERT via coupling it with additional modules. We believe the same strategy could be also applied to other DNN models, which although deliver good overall performance but are too complicated to be modified for a specific problem observed.

In summary, this paper makes the following contributions: (1) We propose a novel framework to supplement BERT for tackling the challenges of *Negation*, *All-of-the-above,* and *None-of-the-above* question types. (2) Experimental results show the proposed approach effectively improves the performance, and thus demonstrate the feasibility of supplementing BERT with additional modules to fix a specific problem observed.

## II. PROPOSED APPROACHES

### A. Problem Formulation

Given a social study problem $Q$ and its corresponding supporting evidence $SE$, our goal is to find the most likely answer from the given candidate-set $A = \{A_1, A_2, \ldots A_n\}$, where $n$ is the total number of available choices/candidates, and $A_i$ denotes $i$-th answer-candidate. This task is formulated as follows, where $\widehat{Ans}$ is the answer to be picked up.

$$\widehat{Ans} = \underset{i=1\ldots n}{\mathrm{argmax}}\, P(A_i|Q, SE, A)$$

### B. Proposed Approach

The system architecture of the proposed framework is shown in Fig 1. It consists of three main components: (1) **Pre-processor**, which turns the answer candidate, "以上皆是" ("All-of-the-above") and "以上皆非" ("None-of-the-above"), into the concatenation of other options (associated with the same question) before inputting the question-choice-evidence combination into the BERT model. Take the last row in Table II as an example, we will replace the original last choice "以上皆非" with "老人^小孩^青壯年". (2) **BERT** model, which is a fine-tuned BERT multiple-choice prediction model described in section III.B. (3) **Answer-Selector**, which is a candidate re-selector that will pick the answer candidate with the lowest score, instead of the one with the highest score (as for other question types), for the negation type question. A given question is classified as negative-type if it includes a negation word within a pre-specified negation word list, which is obtained from the CSSMC training data, and currently consists of {"不會 (will not)", "不能 (cannot)", "不得 (not allow)", "不是 (is not)", "不應該 (should not)", "不可能 (unlikely)", "不需 (do not need)", "不必(do not need)", "不用(do not need)", "沒有 (without)"}.

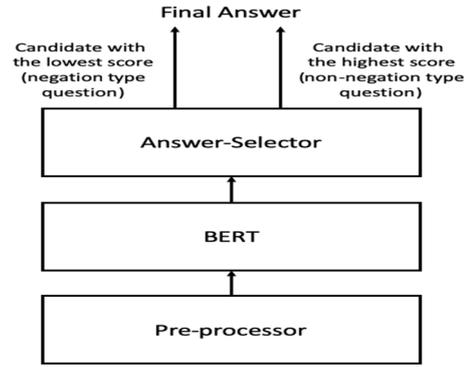

Fig. 1. The system architecture of the proposed framework

TABLE III. CSSMC CORPUS STATISTICS

|  | *Train* | *Dev* | *Test* |
|---|---|---|---|
| #Lessons | 202 | 27 | 26 |
| #Questions | 3,879 | 780 | 778 |
| #Averaged paragraphs/lesson | 11.28 | 13.93 | 10.93 |
| #Averaged sentences/ Lesson | 46.40 | 52.67 | 46.33 |

TABLE IV. CSSMC GOLD-SE1 SUB-SET STATISTICS

|  | *Train* | *Dev* | *Test* |
|---|---|---|---|
| #Lessons | 196 | 27 | 26 |
| #Questions ( #NEG[a] ) ( #AllAbvNonAbv[b] ) | 3,135 (53) (332) | 604 (14) (69) | 563 (15) (56) |
| #Averaged paragraphs/lesson | 11.35 | 13.93 | 10.85 |
| #Averaged sentences/ Lesson | 46.72 | 52.67 | 46.15 |

[a] #NEG: number of Negation-type questions.
[b] #AllNonAbv: number of AllAbvNoneAbv-type questions.

## III. EXPERIMENTS

### A. Data-set

We conduct experiments on the CSSMC data-set. This data-set is collected from three leading publishing houses in Taiwan. There are 255 lessons and 5,437 questions. We randomly divide it into the training, the development, and the test sets. Table III shows the associated statistics.

### B. Baseline

We adopt the BERT [6] model fine-tuned for the multiple-choice task as our baseline, as it is the most widely adopted state-of-the-art model [12]). It is built by exporting the BERT's final hidden layer into a linear layer and then taking a softmax operation. The BERT's input sequence consists of "[CLS] SE [SEP] Q [SEP] $A_i$ [SEP]", where [CLS] and [SEP] are special tokens to represent the classification and the passage separator, respectively, as that defined in [6]. Fig.2 illustrates the architecture of the BERT baseline model.

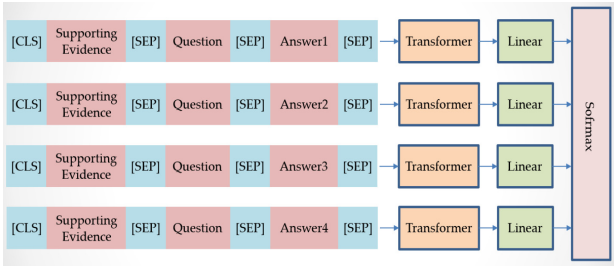

Fig. 2. The architecture of the BERT baseline model

### C. Supporting Evidence (SE) Retrieval

SE is the corresponding shortest passage based on which the system can answer the given question. We hire some outside workers to annotate the associated SE for each question in the CSSMC data-set. For each given question, they first locate its most related lesson from the whole data-set, and then identify the most relevant passage within that lesson. However, considerable questions involve common-sense reasoning, and cannot find their corresponding SEs from the retrieved lesson. We denote the set of questions whose SE can be found in the retrieved lesson as **SE1**, and the set of remaining questions as **SE2**. Only the questions in **SE1** are annotated with their gold SEs. Table IV shows the associated statistics for **SE1**.

### D. Experiments

We conduct two sets of experiments on CSSMC data-set: the first one bases on SE1 sub-set with gold SEs, which is designed to compare the QA *component* performances of different models; and the second one bases on the whole data-set with all SEs directly retrieved from *Lucene* search engine, which is designed to compare different approaches under the real situation. Each set will cover four different models: "***Base***", "***+Neg***", "***+AllAbv&NonAbv***" and "***+Neg+AllAbv&NonAbv***", where "***Base***" is the baseline model, "***Neg***" and "***AllAbv&NonAbv***" denote the cases of adding additional *Answer-Selector* and *Pre-Processor* modules for "*Negation*" and "*All-of-the-above/None-of-the-above*" types, respectively. All experiments trained with the following hyper-parameters: (1) maximum sequence length is 300; (2) learning rate is 5e-5. Table V compares the accuracy rates of various approaches (We report the test-set performance when the best dev-set performance is obtained).

#### 1) SE1 sub-set with gold supporting evidences

In this scenario, we aim to assess the performance of the four different models on SE1 sub-set with gold SEs. This scenario compares the QA component performances of different models. The ***GSE1*** column in Table V gives the associated test-set accuracy rates of various approaches. The "***+Neg***" and "***+AllAbv&NonAbv***" models improves the ***Base*** performance 2.1% (= 87.0% - 84.9%) and 3.0% (= 87.9% - 84.9%) on the test-set, respectively.

To explore the individual effects of the proposed approaches, we conduct two additional experiments on two sub-sets of GSE1 that contains "***Neg-type*** only" and "***AllAbv&NonAve*** only" questions. The ***GSE1-Neg*** and ***GSE1-AllAbv&NonAve*** columns in Table V clearly show the proposed ***Pre-Processor*** and ***Answer-Selector*** modules effectively enhance BERT on these two sub-sets (from 20% to 40%, and from 64.3% to 83.9%, respectively).

TABLE V. TEST-SET PERFORMANCE COMPARISON

| | GSE1[a] | GSE1-Neg[b] | GSE1-AllAbvNonAbv[c] | LSE[d] |
|---|---|---|---|---|
| ***Base*** | 0.849 | 0.200 | 0.643 | 0.692 |
| + *Neg* | 0.870 | **0.400** | - | 0.695 |
| + *AllAbv&NonAbv* | 0.879 | - | **0.839** | 0.719 |
| + *Neg* + *AllAbv&NonAbv* | **0.879** | - | - | **0.725** |

[a] GSE1: SE1 sub-set with gold SEs.
[b] GSE1-Neg: Only negation-type questions within GSE1.
[c] GSE1-AllAbvNonAbv: Only AllAbv&NoneAbv-type questions within GSE1.
[d] LSE: SE1+SE2 with all SEs retrieved from Lucene search engine.

TABLE VI. AN ERROR CASE OF *BASE*+*NEG* ON GSE1-NEG SUB-SET

| Example |
|---|
| **SEs**: "另外，隨著商業興盛，在府城、鹿港、艋舺等大城市，也出現由商人組成的「郊」。「郊」類似現代同業公會，成員除了經營貿易外，也積極參與地方的公共事務。" |
| **Question**: "清朝統治臺灣時期，怎樣的人應該比較<u>沒有</u>共同的血緣？" |
| **Options**: (1)參加同一個宗親會 (2)參加同一個祭祀公業 (3)參加同一個「郊」(4) 在同一座宗祠祭祀祖先 |

TABLE VII. AN ERROR CASE OF *BASE*+ALLABV&NONABV ON GSE1-ALLABV&NONABV SUB-SET

| Example |
|---|
| **SE**: "工業生產如果沒有適當處理，很容易破壞周遭環境，造成空氣汙染、噪音汙染、水質汙染、土地汙染等。例如：工業廢水或是家庭汙水直接排入河流，不僅危害河流生態，有毒物質如果流入大海，通過食物鏈進入人體，更會嚴重損害健康。" |
| **Question**: "志忠家附近有一間工廠，時常將未經處理的汙水排入河川中，這樣可能會造成什麼後果？" |
| **Options**: (1)空氣汙染 (2)噪音 (3)水質汙染 (4)以上皆是 |

The remaining errors of GSE1-NEG and GSE1-AllAbv&NonAbv are mainly due to requiring further inference capability. Table VI shows we need to know "商人 (businessmen)" are people without "共同的血緣 (common consanguinity)". Similar, Table VII shows we need to know "未經處理的汙水排入河川 (untreated sewage discharged into the river)" will cause "水質汙染 (water pollution)".

#### 2) SE1+SE2 with all SEs retrieved from Lucene

Since the gold SE will not be available for real applications, this scenario compares the system performances of different models under the real situation. That is, we test various models with all the SEs retrieved from a search engine (i.e., *Apache Lucene*). Furthermore, to support those questions that cannot find their associated SES from the lessons (i.e., ***SE2*** sub-set), we introduce *Wikipedia* as an external knowledge resource for providing possible SEs. We first let *Apache Lucene* search the textbook and Wikipedia separately to get two possible SEs. Afterwards, we construct a fused SE via concatenating these two SEs with the format "Textbook-SE [SEP] Wiki-SE" (where "Textbook-SE" and "Wiki-SE" denote the two SEs retrieved from the Textbook and Wikipedia, respectively).

Experiment results (the ***LSE*** column in Table V) show that both the ***Pre-processor*** and the ***Answer-Selector*** could effectively supplement BERT; and it could further improve the performance (about 3.3% = 72.5% - 69.2%) when they are jointly adopted. Furthermore, the accuracy of the ***Base*** model on LSE is only 69.2%, which is significantly lower than that of GSE1 (i.e., 84.9%). It thus clearly illustrates that extracting good SE is essential in QA tasks. Last, to show the influence of incorporating Wikipedia, we conduct another experiment to let *Apache Lucene* search the textbook only. The "*+Neg+AllAbv&NonAbv*" model now drops to 70.4% (not shown in Table V) from 72.5%. The result shows that Wikipedia does provide required common-sense for some cases.

TABLE VIII. EXAMPLES OF DIFFERENT ERROR TYPES

| Error Type | Questions |
|---|---|
| Incorrect supporting evidence (52%) | **Wrong SE**: "*清朝統治臺灣初期，漢人渡海來臺後，往往同鄉人聚居在一起，並且建築廟宇供奉共同信仰的神明。*"<br>**Question**: "*臺灣有許多從中國移民來的漢人，來臺要渡過危險的臺灣海峽，所以什麼神明就被所有移民所共同信仰？*"<br>**Options**: (1)關公 (2)土地公 (3)媽祖 (4)三山國王 |
| Requiring advanced inference capability (48%) | **SE**: "*刑法對傷害他人的行為加以處罰；民法則以損害賠償的方式，請問牛奶的保存期限過了沒？（相關法律：民法、消費者保護法、食品安全衛生管理法）*"<br>**Question**: "*小花在超市買到過期的餅乾，請問該超市的販售行為違反什麼法律？*"<br>**Options**: (1)刑法 (2)憲法 (3)教育基本法 (4)食品安全衛生管理法 |

## IV. ERROR ANALYSIS AND DISCUSSION

We randomly select 40 error cases from the test set of the "*+Neg+AllAbv&NonAbv*" model under the scenario "*SE1+SE2 with all SEs retrieved from Lucene*". We found all of them come from two sources: (1) not retrieving correct support evidence (52%), and (2) requiring deep inference (48%). Table VIII shows an example for each category.

For the first example, the retrieved SE is irrelevant to the question; our model thus fails to get the correct answer. On the other hand, the second example illustrates that the model requires further inference capability to know "牛奶的保存期限過了沒 (Has the milk had expired ?)" and "在超市買到過期的餅乾 (I bought expired cookies in the supermarket.)" are similar events related to "食品安全衛生管理法 (Act Governing Food Safety and Sanitation)".

## V. RELATED WORK

Before 2015, most previous works of entailment judgment adopt statistical approaches [2-3]. Afterwards, the neural network models are widely adopted due to the availability of large data-sets (such as RACE [13] and SNLI [14]). Parikh et al. [4] presented the first alignment-and-attention mechanism to achieve SOTA results on SNLI data-set. Chen et al. [5] further proposed a sequential inference model, based on the chain LSTMs, which outperforms previous models. Recently, several pre-trained language models (such as BERT [6], XLNET [7], RoBERTa [8] and ALBERT [9]) have obtained superior performance on considerable *MC RC* tasks. However, they achieve the remarkable results mainly via utilizing the surface features [15].

Instead of adopting a new inference NN, our proposed approach mainly supplements BERT with additional modules to address two specific problems that are not well handled.

## VI. CONCLUSION

We present a novel framework to supplement BERT with additional modules to address two specific problems (i.e., *Negation*, *All-of-the-above,* and *None-of-the-above*) that BERT handles poorly. The proposed approach suggests a new way to enhance a complicated DNN model with additional modules to pinpoint some specific problems found in error analysis. Experimental results show the proposed approach effectively improves the performance, and thus demonstrate the feasibility of supplementing BERT with additional modules to fix specific problems observed.